\documentclass{article}
\usepackage{spconf,amsmath,graphicx}

\title{Multiple View Generation and Classification of Mid-wave Infrared Images using Deep Learning}
\name{Maliha Arif , Abhijit Mahalanobis}
\address{Center for Research in Computer Vision,
University of Central Florida\\
4000 Central Florida Blvd, 32816, Orlando, FL, USA}

\begin{document}
%
\maketitle
\begin{abstract}
We propose a novel study of generating unseen arbitrary viewpoints for infrared imagery in the non-linear feature subspace .  Current methods use synthetic images and often result in blurry and distorted outputs. Our approach on the contrary understands the semantic information in natural images and encapsulates it such that our predicted unseen views possess good 3D representations. We further explore the non-linear feature subspace and conclude that our network does not operate in the Euclidean subspace but rather in the Riemannian subspace. It does not learn the geometric transformation for predicting the position of the pixel in the new image but rather learns the manifold. To this end, we use t-SNE visualisations to conduct a detailed analysis of our network and perform classification of generated images as a low-shot learning task. 

%
\end{abstract}
\section{Introduction}
\label{sec:intro}

View synthesis has been a long-standing problem with works such as \cite{zhou2016view,tatarchenko2016multi} notable in giving good qualitative solutions. However,majority of these works \cite{ji2017deep,dosovitskiy2015learning} involve the use synthetic data which is unlimited in quantity and can be created with desired azimuth, elevation, degree variation and background setting. ShapeNET \cite{chang2015shapenet} dataset which consists of over 8000 synthetic car and chair models and SURREAL dataset \cite{jin2018learning} which contains synthetic human motion sequences are two such widely used datasets.  We make the problem challenging by dealing with natural infrared images of MWIR (medium wave) frequency in the 'DSIAC-ATR Image database' . These images are inherently complex and carry intrinsic details such as background , clutter , moving targets, day and night variations which are difficult to comprehend and predict.Collection of such data is also a big hassle due to which we show that our generated images can also be used in the training corpus for various image processing tasks including classification. Since infrared data is extremely scarce, our study assists in creating data as close to real world natural images.
Most of the related view generation studies have been categorized in two categories. Geometry based where the weights of the neural network learn the geometric transformation of a pixel between two views, the input and output view. Second, where view prediction is seen as a learning problem and the network is trained to understand the 3D representation of objects. Few works have exploited the use of both approaches \cite{zhou2016view} to come up with hybrid methods of using geometry as well as learning for view morphing. We however adopt the learning based approach and assert via our results that our approach is not just learning the 3D representation of objects but rather its manifold. We believe the pixels in the input and output view are highly correlated and if observed on a manifold lie at the same point. We make use of this assumption and do novel view synthesis as a manifold learning task and not as a geometry based task. 
 We do not propose to explicitly learn the 3D information of our model but rather we want to exploit the latent space embedding of autoencoders to learn the underlying semantic information and pixel relation between input and novel output views. We show via experiments that our network learns the underlying feature subspace in the non-linear domain.  Our approach is entirely unsupervised and we can say that while learning the manifold, our network also learns to classify objects and becomes a classification model as suggested in \cite{tatarchenko2019single}.
We are able to generate novel views both during day and night, which is significant in infrared imagery.
We make following contributions:
\begin{itemize}
    \item We present a novel study of generating unseen views for infrared imagery.
    \vspace{-0.3cm}
    \item We propose a method that creates realistic looking training images to assist in low-shot learning and classification.
    \vspace{-0.3cm}
    \item We perform experiments that show view generation is a manifold learning task that exploits latent space embeddings.
\end{itemize}

\begin{figure}
\includegraphics[width= 1.0\linewidth,height= 1.1\linewidth]{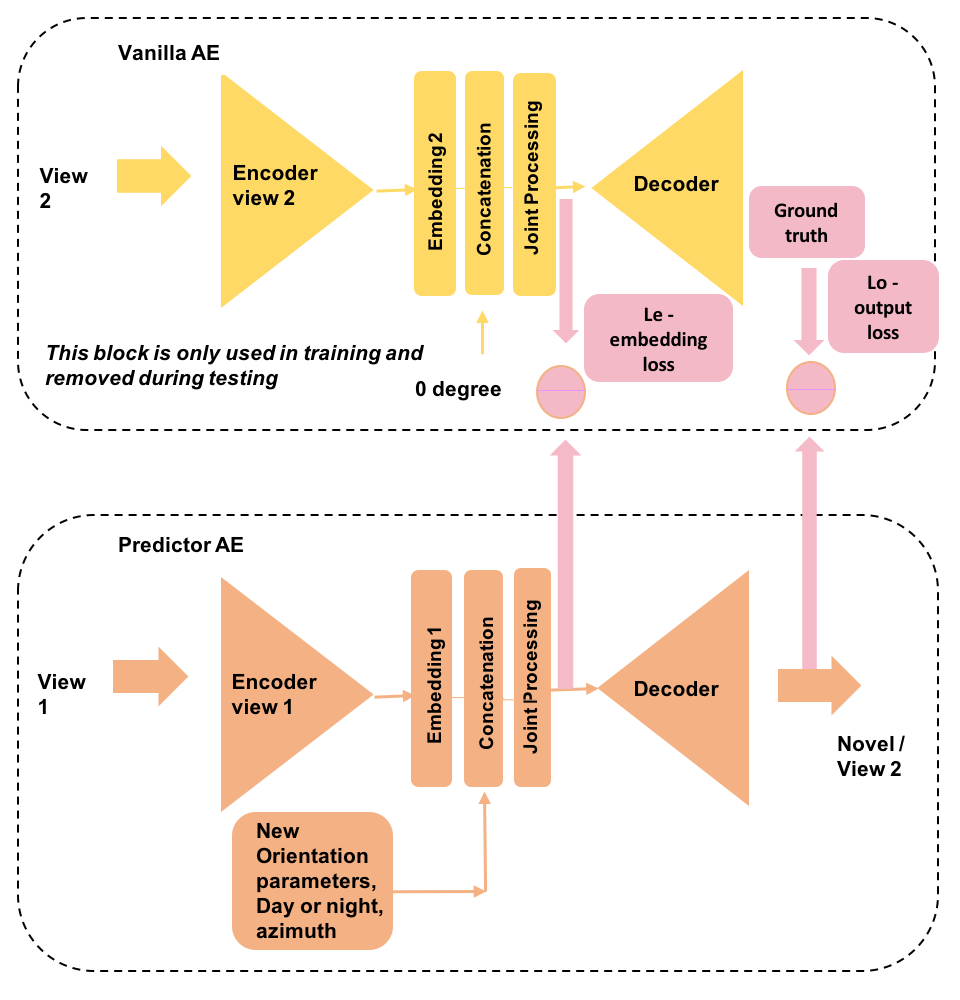}
\caption{Our proposed network framework and training technique.We compute loss between our embeddings and final generated view using mean square error.}
\label{block_diagram}
\end{figure}

\vspace{-0.2cm}

\section{Proposed Approach}
\label{sec:method}

\begin{figure}
\includegraphics[width= 1.0\linewidth,height= 0.8\linewidth]{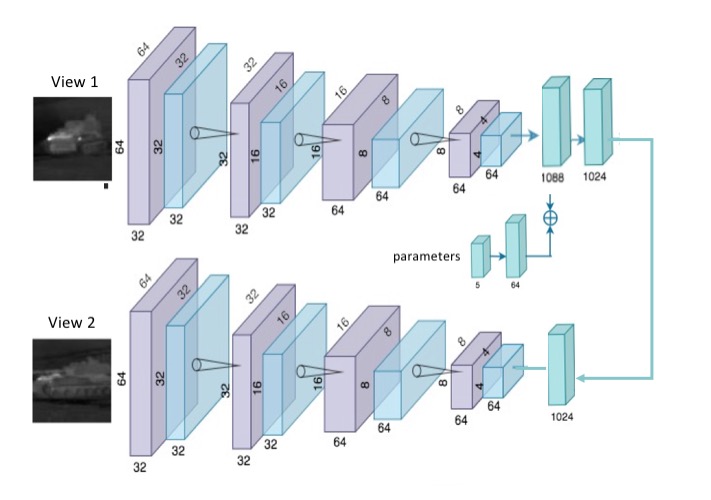}
\caption{Our implemented network architecture. For both our vanilla and predictor autoencoder, we use 4 convolutional and 4 deconvolutional layers. We use input image size of 64 x 64.}
\label{architecture2-edited}
\end{figure}

\subsection{Deep Convolutional Neural Network based Autoencoder framework}
\label{sec:Architecture}
Feature extraction and feature selection(a NP hard problem) when done manually pose a lot of problems. Are the features good or variable enough to assist in network learning are questions that need addressing. Autoencoders have emerged as an alternative to feature extraction and selection in the non-linear feature subspace \cite{charte2018practical}. We propose a deep convolutional neural network based autoencoder with emphasis on exploiting its latent space and learning the manifold. 
Our proposed method is illustrated in Figure \ref{block_diagram}.Block 1 denotes a vanilla autoencoder \cite{berthelot2018understanding} which acts as a guide and paves way to formulate the first term of our loss function. This block is trained separately and is used to supply the ground truth embeddings for the desired viewpoints.Block 2 denotes the predictor network which is an exact replica of the first block except that it receives an input view and produces an unseen output view. We adapt the network proposed in \cite{tatarchenko2016multi} to create the autoencoder blocks.
  The autoencoders have 4 convolutional blocks in encoder and decoder with consecutive layers having 32, 32,64 and 64 filters respectively.The filter size is $5 \times 5$ for the first layer and then $3 \times 3$ for the following layers.We use an image size of $64 \times 64$ for our network. We tried using image size of $128 \times 128$ as well but obtained better results with the former. The encoder's output is $4 \times 4 \times 64$ dimensional which we flatten to create a 1024 long embedding representing the input view. We introduce feature fusion in the network and pass a 5 dimensional long vector through a 64 dimensional fully connected layer to be processed independently.In an unsupervised setting, the 1024 embedding of our input view which is of varying orientation, azimuth and time of day is concatenated with our independent feature vector. This latent embedding which is now 1088 dimensional long is jointly processed using two further fully connected layers as shown in the detailed architecture in Figure \ref{architecture2-edited}. We use leaky relu activation.For the last layer, we use hyperbolic tangent since it produces steeper slope and stronger gradients as stated by LeCun et al. \cite{charte2018practical,orr2003neural}.

\subsection{Objective function}
\label{sec:lossfunction}
Our objective function is a sum of two separate loss terms. The first term aims to guide the latent space embedding to a set of values which best describe our desired viewpoint. This also means, we must have the ground truth for the embeddings. Section \ref{sec:Architecture} describes how we obtain these ground truth embeddings. Once we obtain them, we compute our first term in the loss function $L_e$ . 
\begin{equation}
L_e = \sum \lVert e_{2} - e_{1}\rVert^{2}
\end{equation}
Here $e_2$ is the ideal embedding which we generate using our vanilla autoencoder and $e_1$ is the embedding which we obtain while training our predictor network. This is depicted in Figure \ref{block_diagram}.We use mean square error to calculate the loss. 
The second term in our objective function aims to reduce the loss between our predicted output view and its ground truth. It is given by :
\begin{equation}
L_{o} = \sum\lVert y_{2} - y_{1}\rVert^{2}
\end{equation}
where $y_2$ and $y_1$ are the ground truth and generated views respectively. 
We use $L_e$ and $L_o$ to compute our final reconstruction loss  $L_t$ which is a sum of these two error terms as observed below. 
\begin{equation}
    L_{t} = L_{e} + L_{o}
\end{equation}
\begin{equation}
    L_t = \sum \lVert e_{2} - e_{1}\rVert^{2} + \sum\lVert y_{2} - y_{1}\rVert^{2}
\end{equation}

\begin{figure}
\includegraphics[width= 1.0\linewidth,height= 1.0\linewidth]{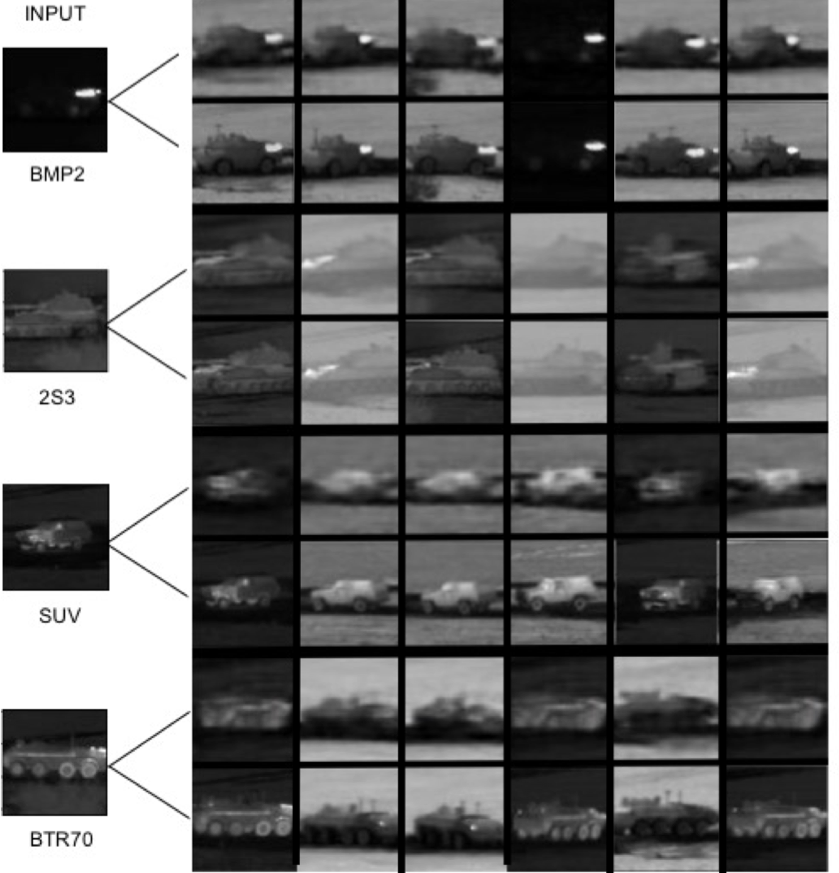}
\caption{Qualitative results of our method. The left side of the figure shows the single input image which is used as View 1 for our network. Top row of the images on the right are the predicted images generated using the single input image, the bottom row shows their corresponding ground truth views. Results for other classes have been shared in supplementary material.}
\label{Results-final}
\end{figure}

\section{Experiments}
\label{sec:pagestyle}


\subsection{Dataset}
We work on ‘DSIAC-ATR Image Dataset’. The Automated Target Recognition (ATR) Algorithm Development Image database \cite{ATRdataset} has been collected by US Army Night Vision and Electronic Sensors Directorate (NVESD) with the intent of supporting work in the infrared and ATR domain. This database contains MWIR (mid-wave infrared) images of military and civilian vehicles with 10 non-human categories. Each category has videos collected during the day and night at different ranges (distance from the camera capturing the target). Each video shows the object moving in two 360 degree circles. The ranges lie between 900-5000 m. This means that all of the videos have been captured from a very far distance which makes our task of view prediction extremely difficult and painstaking.Most of the space in the images is taken up by background and clutter with the object of interest appearing very small and distorted. The 10 object categories consist of SUVs, pickup trucks, battle tanks and anti-aircraft missile launchers.Since these are infrared images; all vehicles are grayscale and show uneven heat distribution. Our work as a result is very different from other view prediction research studies   \cite{tatarchenko2016multi,dosovitskiy2015learning,zhou2016view,jin2018learning,ji2017deep,kulkarni2015deep} which use evenly distributed colored objects, have unlimited training samples and also have the ability to create their own azimuth, elevation and range parameters. We on the contrary are dealing with annotations supplied with the dataset. Hence parameters like range, azimuth, elevation , light intensity are all beyond our control.

\begin{table}
\begin{center}
\setlength{\tabcolsep}{0.5pt}
\renewcommand{\arraystretch}{1.0}
\small
\begin{tabular}{c|c|c|c}
\hline
		  \textbf{Model}    & \textbf{Input } &\textbf{Conv layers} & \textbf{Average}\\
		           &\textbf{dimensions}  & \textbf{encoder,decoder}   & \textbf{error} \\   
\hline	         
        MV3D\cite{tatarchenko2016multi} & $128 \times 128$ & 5,5   & 0.0056   \\
\hline
Ours +     & $64 \times 64$ & 4,4   & 0.0011\\
MSE loss   &           &               &  \\
\hline
Ours +     & $64 \times 64$ & 4,4    & 0.0009\\
New loss   &           &               &  \\
\hline
\end{tabular}
\end{center}

\caption{
A quantitative comparison - average error with different methods,input dimensions and objective functions.
}
\label{table_quant_comparisons}
\end{table}

\subsection{Training and Implementation details}
We first convert the videos from arf format to avi format which are then converted to $640 \times 580$ dimensional images \cite{arif2020view}. These images are used to obtain the objects of interest which are cropped to a size of $64 \times 64$ removing background as much as possible. The dataset has 5 ranges, we work with objects that are at a range of 1000 m or less so that the object is visible. This threshold gives us approximately 180 images per object of different azimuth. We create our dataset of views which are all 5 degrees apart and contain images both from day and night time.These are used to create training pairs with varying input and output viewpoints.We end with 30k training pairs and 10k test pairs.We use Adam optimizer \cite{kingma2014adam} and adaptive learning rate methods. We use learning rate of 1e-3 initially and decrease it to 1e-4 for the last few epochs. We reach convergence after 80 epochs.We extract ground truth embeddings from block 1. These embeddings are then fed with the ground truth images for view 2 in block 2 of our proposed method. Our predictor network uses our new loss function as defined in section \ref{sec:lossfunction} to compute the final reconstruction error given in Table \ref{table_quant_comparisons}. We perform multiple experiments; baseline based on \cite{tatarchenko2016multi,dosovitskiy2015learning,kulkarni2015deep,zhou2016view}, our new architecture with just simple mean square error loss and then with our new objective function. We use a batch size of 64 for all our experiments and keras with tensorflow backend for implementation. Figure \ref{Results-final} illustrates our obtained qualitative results.

\section{Evaluation}
\label{sec:typestyle}

\subsection{Classification and Low-shot learning}
\begin{figure}
\includegraphics[width= 1.0\linewidth,height= 0.8\linewidth]{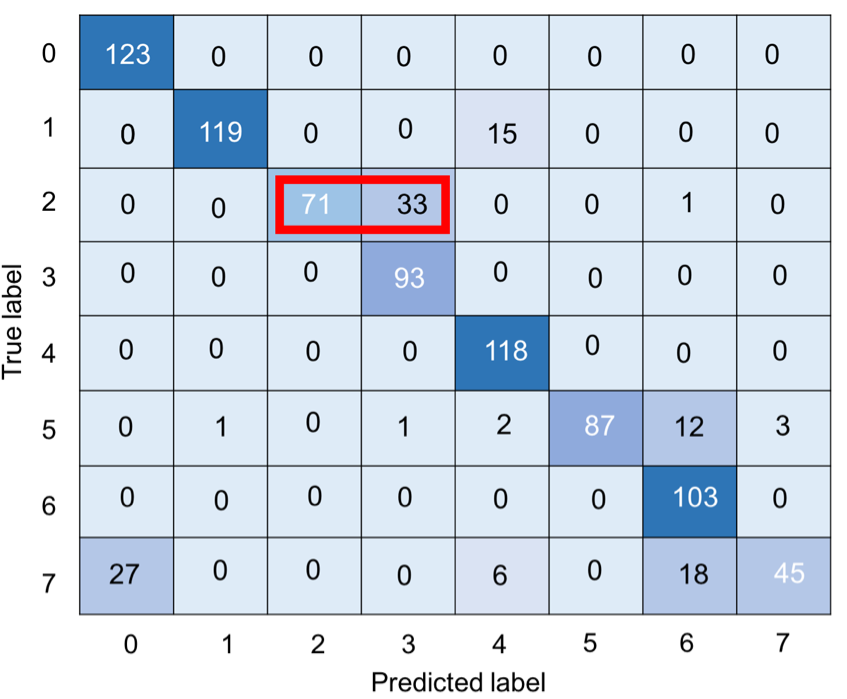}
\caption{The confusion matrix shows correct predictions for Class 2 'BRDM2'-an infantry vehicle when we use synthesized images for training.}
\label{t-confusion_matrix}
\end{figure}

Studies have shown that it is difficult to judge the generated images both qualitatively and quantitatively . Visual analysis depends on how the reader observes and assess the image. A good evaluation metric in terms of quality is their classification. This has given rise to the concepts of low-shot \cite{wang2018low}, zero-shot and one-shot learning.These techniques are becoming popular because it is difficult to collect and annotate real world datasets.We perform such a classification experiment using 8 classes and replace class 2- BRDM2’s training images with only our network generated images. We have already shown in Figure \ref{Results-final} that our network is capable of generating multiple novel views using a single image. We use a pretrained VGG-16 \cite{simonyan2014very} network and place our 8-class classifier on top. We test using only real-world images for all classes even BRDM2. We obtain 68\% classification accuracy with very little mispredictions for class 2- BRDM2 shown in Figure \ref{t-confusion_matrix} as compared to a 79\% accuracy when we use real images for training affirming that our generated images are of good quality and can be used for experimental research. 

\subsection{Manifold learning}
\begin{figure}
\includegraphics[width= 1.0\linewidth,height= 0.6\linewidth]{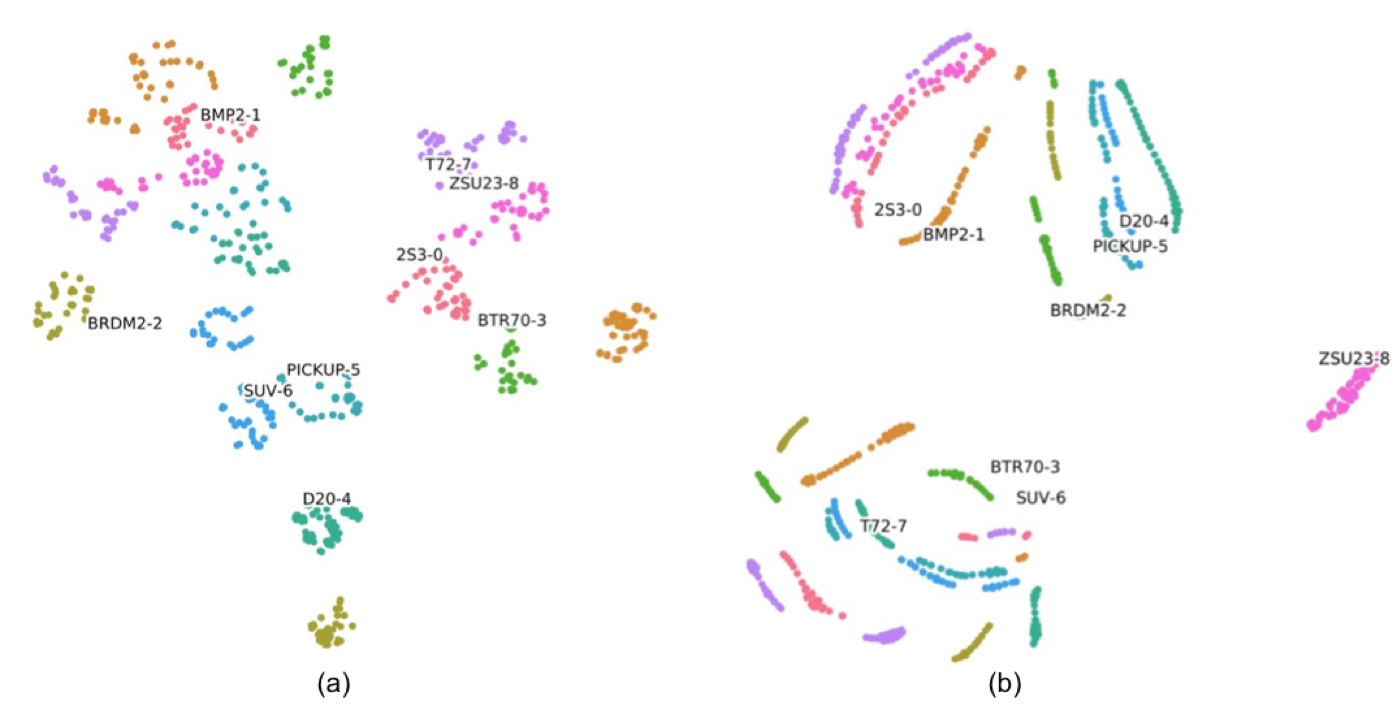}
\caption{t-SNE visualizations for the two embeddings.Object classes and their full names have been shared in the supplementary material. }
\label{t-SNE1}
\end{figure}

Manifold representations are not only convenient for visualizing high dimensional data, but also provide insights about how different classes inter-relate. A lower dimensional embedding of the manifold ensures that the geodesic relations between the data points is preserved. The problem of predicting new views from a few original views may be viewed as learning the behavior of the manifold, and the transition from one view to another as a movement along its surface. Thus, we interpret our results in terms of how the original and predicted views behave in the latent space. We verify that the predicted views and the ground truth images are “close” on the underlying manifold. The t-SNE\cite{maaten2008visualizing} method models each high-dimensional object by a two-or three-dimensional point in such a way that similar objects are modeled by nearby points and dissimilar objects are modeled by distant points with high probability.  We visualize two embeddings from our trained network, one obtained from the output of the encoder, second obtained after we inject our orientation parameters and process the concatenated features. Figure \ref{t-SNE1} (a) shows our embeddings before feature fusion. We can see different object classes, for each object class, we observe two clusters, one showing the night and the other, the day view. The data points however seem scattered and do not show uniformity. Figure \ref{t-SNE1} (b) shows the results for our embeddings after feature fusion and what the network will predict given the unseen viewpoints. It clearly depicts uniformity and precise clustering which affirms that the network learnt the manifold of the objects.

\vspace{-0.4cm}
\section{Conclusion}
We have performed a novel study of predicting unseen views on infrared imagery. The use of deep learning and a CNN based autoencoder for exploiting the latent space embeddings and predict novel views has shown promising results. Our proposed method has provided us an effective modeling of entities present in our dataset. The method is able to understand the difference in object shape and class without any reliance on color. We are able to show that the network learns the manifold and operates in riemannian feature space. Moreover our experimental evaluations show the quality of our generated images and the effectiveness of our technique.

\textbf{Acknowledgement}: The authors gratefully recognize the support of Leonardo DRS for funding parts of this research.




%

%
%


\bibliographystyle{IEEEbib}
\bibliography{strings,refs}

\begin{thebibliography}{10}

\bibitem{zhou2016view}
Tinghui Zhou, Shubham Tulsiani, Weilun Sun, Jitendra Malik, and Alexei~A Efros,
\newblock ``View synthesis by appearance flow,''
\newblock in {\em European conference on computer vision}. Springer, 2016, pp.
  286--301.

\bibitem{tatarchenko2016multi}
Maxim Tatarchenko, Alexey Dosovitskiy, and Thomas Brox,
\newblock ``Multi-view 3d models from single images with a convolutional
  network,''
\newblock in {\em European Conference on Computer Vision}. Springer, 2016, pp.
  322--337.

\bibitem{ji2017deep}
Dinghuang Ji, Junghyun Kwon, Max McFarland, and Silvio Savarese,
\newblock ``Deep view morphing,''
\newblock in {\em Proceedings of the IEEE Conference on Computer Vision and
  Pattern Recognition}, 2017, pp. 2155--2163.

\bibitem{dosovitskiy2015learning}
Alexey Dosovitskiy, Jost Tobias~Springenberg, and Thomas Brox,
\newblock ``Learning to generate chairs with convolutional neural networks,''
\newblock in {\em Proceedings of the IEEE Conference on Computer Vision and
  Pattern Recognition}, 2015, pp. 1538--1546.

\bibitem{chang2015shapenet}
Angel~X Chang, Thomas Funkhouser, Leonidas Guibas, Pat Hanrahan, Qixing Huang,
  Zimo Li, Silvio Savarese, Manolis Savva, Shuran Song, Hao Su, et~al.,
\newblock ``Shapenet: An information-rich 3d model repository,''
\newblock {\em arXiv preprint arXiv:1512.03012}, 2015.

\bibitem{jin2018learning}
Shi Jin, Ruiynag Liu, Yu~Ji, Jinwei Ye, and Jingyi Yu,
\newblock ``Learning to dodge a bullet: Concyclic view morphing via deep
  learning,''
\newblock in {\em Proceedings of the European Conference on Computer Vision
  (ECCV)}, 2018, pp. 218--233.

\bibitem{tatarchenko2019single}
Maxim Tatarchenko, Stephan~R Richter, Ren{\'e} Ranftl, Zhuwen Li, Vladlen
  Koltun, and Thomas Brox,
\newblock ``What do single-view 3d reconstruction networks learn?,''
\newblock in {\em Proceedings of the IEEE Conference on Computer Vision and
  Pattern Recognition}, 2019, pp. 3405--3414.

\bibitem{charte2018practical}
David Charte, Francisco Charte, Salvador Garc{\'\i}a, Mar{\'\i}a~J del Jesus,
  and Francisco Herrera,
\newblock ``A practical tutorial on autoencoders for nonlinear feature fusion:
  Taxonomy, models, software and guidelines,''
\newblock {\em Information Fusion}, vol. 44, pp. 78--96, 2018.

\bibitem{berthelot2018understanding}
David Berthelot, Colin Raffel, Aurko Roy, and Ian Goodfellow,
\newblock ``Understanding and improving interpolation in autoencoders via an
  adversarial regularizer,''
\newblock {\em arXiv preprint arXiv:1807.07543}, 2018.

\bibitem{orr2003neural}
Genevieve~B Orr and Klaus-Robert M{\"u}ller,
\newblock {\em Neural networks: tricks of the trade},
\newblock Springer, 2003.

\bibitem{ATRdataset}
US~Army~Night Vis. and Elec. Sensors~Directorate (NVESD),
\newblock ``Atr algorithm development image database,''
\newblock {\em Journal of machine learning research}, pp. 1--10, 2012.

\bibitem{kulkarni2015deep}
Tejas~D Kulkarni, William~F Whitney, Pushmeet Kohli, and Josh Tenenbaum,
\newblock ``Deep convolutional inverse graphics network,''
\newblock in {\em Advances in neural information processing systems}, 2015, pp.
  2539--2547.

\bibitem{arif2020view}
Maliha Arif and Abhijit Mahalanobis,
\newblock ``View prediction using manifold learning in non-linear feature
  subspace,''
\newblock in {\em MIPPR 2019: Pattern Recognition and Computer Vision}.
  International Society for Optics and Photonics, 2020, vol. 11430, p. 114301N.

\bibitem{kingma2014adam}
Diederik~P Kingma and Jimmy Ba,
\newblock ``Adam: A method for stochastic optimization,''
\newblock {\em arXiv preprint arXiv:1412.6980}, 2014.

\bibitem{wang2018low}
Yu-Xiong Wang, Ross Girshick, Martial Hebert, and Bharath Hariharan,
\newblock ``Low-shot learning from imaginary data,''
\newblock in {\em Proceedings of the IEEE Conference on Computer Vision and
  Pattern Recognition}, 2018, pp. 7278--7286.

\bibitem{simonyan2014very}
Karen Simonyan and Andrew Zisserman,
\newblock ``Very deep convolutional networks for large-scale image
  recognition,''
\newblock {\em arXiv preprint arXiv:1409.1556}, 2014.

\bibitem{maaten2008visualizing}
Laurens van~der Maaten and Geoffrey Hinton,
\newblock ``Visualizing data using t-sne,''
\newblock {\em Journal of machine learning research}, vol. 9, no. Nov, pp.
  2579--2605, 2008.

\end{thebibliography}

\end{document}